\pdfoutput=1

\documentclass[11pt]{article}

\usepackage[]{acl}
\usepackage{graphicx}

\usepackage{tabularx} %
\usepackage{booktabs} %
\usepackage{enumitem}
\usepackage{makecell}
\usepackage{multirow}
\usepackage{amsmath}
\usepackage{mdframed}
\usepackage{hyperref}
\usepackage{times}
\usepackage{latexsym}
\usepackage[colorinlistoftodos]{todonotes}
\usepackage[most]{tcolorbox}
\usepackage[T1]{fontenc}

\usepackage[utf8]{inputenc}

\usepackage{microtype}

\usepackage{color-edits}
\addauthor{gn}{magenta}
\addauthor{vijay}{teal}
\addauthor{ritu}{orange}
\addauthor{sw}{blue}
\addauthor{sg}{violet}

\title{ Better Synthetic Data by Retrieving and Transforming \\ Existing Datasets}

\author{Saumya Gandhi\thanks{~~equal contribution.}\hspace{0.45em}, Ritu Gala$^{*}\hspace{0.08em}$, Vijay Viswanathan,\\\textbf{Tongshuang Wu}, \textbf{Graham Neubig }\\ Carnegie Mellon University} %

\begin{document}
\maketitle
\begin{abstract}
Despite recent advances in large language models, building dependable and deployable NLP models typically requires abundant, high-quality training data. However, task-specific data is not available for many use cases, and manually curating task-specific data is labor-intensive. Recent work has studied prompt-driven synthetic data generation using large language models, but these generated datasets tend to lack complexity and diversity. To address these limitations, we introduce a method, \textit{DataTune}, to make better use of existing, publicly available datasets to improve automatic dataset generation. DataTune performs \textit{dataset transformation}, enabling the repurposing of publicly available datasets into a format that is directly aligned with the specific requirements of target tasks. On a diverse set of language-based tasks from the BIG-Bench benchmark, we find that finetuning language models via DataTune improves over a few-shot prompting baseline by 49\% and improves over existing methods that use synthetic or retrieved training data by 34\%. We find that dataset transformation significantly increases the diversity and difficulty of generated data on many tasks. We integrate DataTune into an open-source repository to make this method accessible to the community.\footnote{\href{https://github.com/neulab/prompt2model}{https://github.com/neulab/prompt2model}}

\end{abstract}

\begin{figure}[ht]
\centering
\includegraphics[ width=\linewidth]{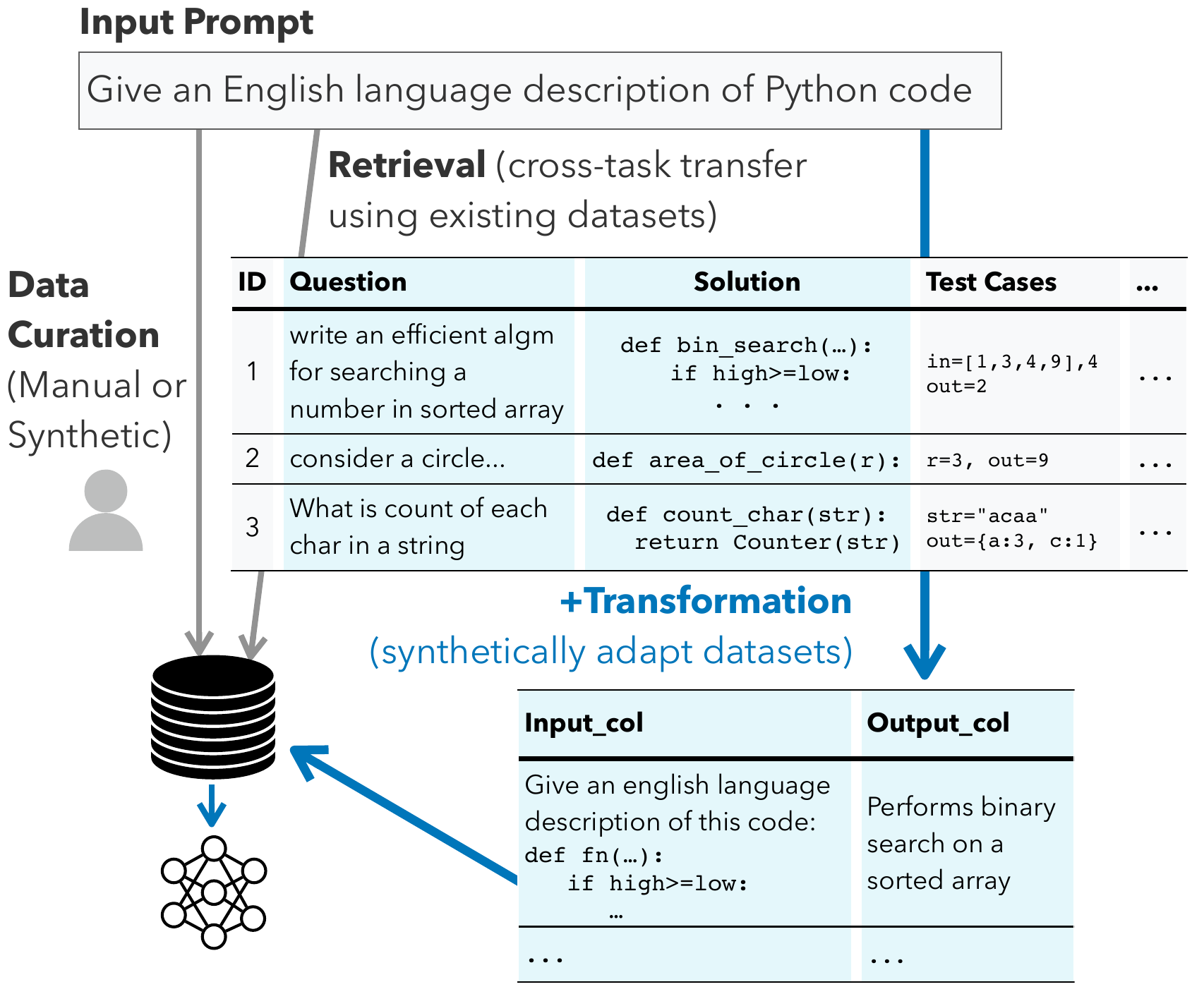}
\caption{Obtaining task-specific annotated data can be tricky. Existing solutions include (1) \textit{data generation methods} either by employing human annotators (incurring high costs) or synthetically, such as using LLMs (risking low diversity) or (2) \textit{cross-task transfer}, where related but task-misaligned datasets are used (for instance, for the task of generating English language descriptions based on code, this could be a public dataset with coding questions, solutions, and test cases but no explicit descriptions). Our approach combines these strategies by adaptively transforming existing datasets for the target task (using the "solution" field from the public dataset and asking an LLM to create description or make any formatting changes required) preserving original dataset diversity while ensuring the quality of synthetically generated data. }
\label{fig:intro-fig}
\end{figure}

\section{Introduction}

The major bottleneck in modern AI research is data. Despite the paradigm-shifting developments of pretraining and prompting, the recipe for achieving peak performance on any particular task has hardly changed: obtain large amounts of high-quality training data and fine-tune your model. This is particularly valuable for developing models with fewer than 3 billion parameters; in this regime, supervised finetuning can be significantly more effective than in-context learning \cite{mosbach-etal-2023-shot}.
However, this recipe can be challenging for specialized or novel tasks where task-specific annotated data is limited.

Prior works have proposed strategies for fine-tuning in this low-resource setting. The most intuitive way is \emph{obtain labeled data} through manual curation~\cite{callison2009fast,zhang2022survey}, which assumes access to domain experts and may require considerable financial resources to compensate annotators fairly \citep{huang-etal-2023-incorporating}. An increasingly popular alternative to large-scale manual annotation is to synthetically produce datasets using existing large models~\cite{wang2022self,peng2023instruction,tang2023does}. We will refer to this method as \textit{synthetic data generation}. Unfortunately, directly generating synthetic datasets that simultaneously have high correctness and sufficient diversity is difficult. Models trained on synthetic datasets are typically significantly worse than those trained on manually curated data when standardizing for dataset size, suggesting that current synthetic dataset generation methods still leave significant room for improvement in dataset quality \citep{ding-etal-2023-gpt}.

On the other hand, \emph{cross-task transfer} entirely sidesteps the need for strictly in-domain data by training models either in a multi-task fashion on a wide range of datasets~\citep{sanh2021multitask} or on task-specific datasets closest to the new target task~\citep{vu2020exploring}.
We focus on the latter, which we refer to as \textit{existing data}. Recent methods such as Prompt2Model \cite{prompt2model} combine both of the above methods, showing the additive benefits of synthetic data generation and existing datasets for finetuning in few-shot settings. However, they report that even for the machine reading question answering dataset of SQuAD \citep{rajpurkar-etal-2016-squad}, a dataset considered today to be largely ``solved'', training on \textit{existing data} alone is significantly worse than training on manually or synthetically created data.

To overcome the limitations of these approaches, we introduce DataTune, a system that automatically repurposes public datasets for new tasks. DataTune identifies the most relevant public datasets and uses a large language model to transform them into a format aligned with the target task's needs. For instance, for a task requiring descriptions of Python code in English (as shown in Figure \ref{fig:intro-fig}), DataTune finds a dataset with programming questions and code solutions (along with other, irrelevant data columns). It transforms the retrieved dataset by using the code solutions as input and generating a synthetic description of the code as output. This approach maintains the original dataset's diversity while matching the task specification, boosting performance by 22 points over baselines. We refer to this synthetic adaptation of publicly available datasets as \textit{Dataset Transformation}.

We evaluate the effectiveness of DataTune on six challenging language-based tasks from the BIG-Bench benchmark \citep{srivastava2023beyond}, which are designed to gauge the system's performance across diverse NLP task categories.
When compared to few-shot prompting of the same base model (\textit{Mistral-7B}) without fine-tuning, DataTune-enhanced models improve by an average of 5.2 points on five tasks, demonstrating its value for domain-specific fine-tuning. DataTune also can be used additively with existing synthetic dataset generation approaches, yielding an 8-point improvement over the few-shot prompting baseline. Comparing DataTune with existing methods of synthetic data generation, we find that DataTune often produces more difficult and diverse examples, and on a small sample of data we observe that these benefits do not come at the expense of data correctness.

\section{Problem Setup}
We study the problem of how to automatically \textit{retrieve} and \textit{transform} existing datasets to prepare a fine-tuning dataset for a new task. In our problem setting, a user specifies the task of interest with a textual prompt $T$ (optionally containing a few demonstration examples). We assume access to a large, diverse collection of labeled datasets $\mathcal{D}$, a large language model that can be prompted, and a small language model $M$ that we can fine-tune.

The goal here is to automatically generate a synthetic training dataset $D'$ which, after finetuning $M$ on $D'$, will improve $M$'s ability to satisfy the task specification. For each task $T$ in a known set of task descriptions $\mathcal{T}$, we can measure our progress towards this goal by evaluating the trained model $M_T$ against labeled data obtained for task $T$.

\begin{figure*}[t]
    \includegraphics[width=\textwidth]{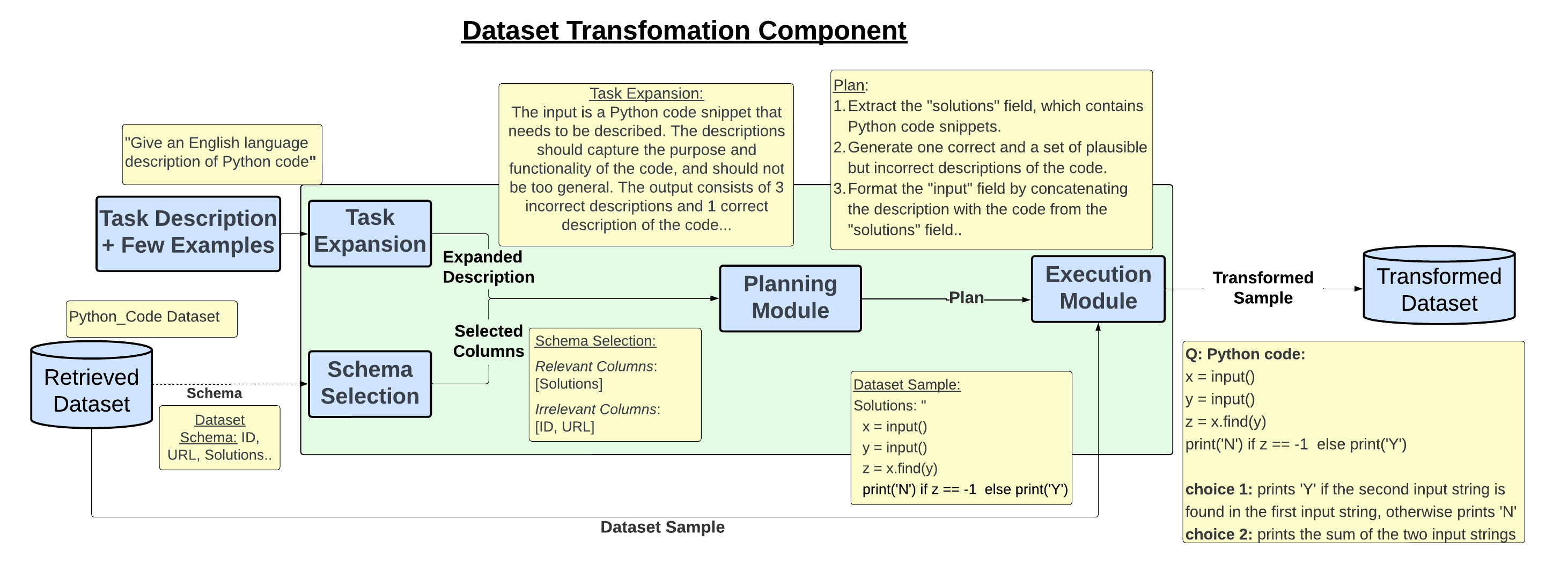}
    \caption{The data transformation component of DataTune, explained with an example (in yellow). }
    \label{fig:data_transformation_architecture}
\end{figure*}
\section{Methods}

DataTune focuses on selecting and transforming a dataset to align it with a specific task.
First, it finds relevant datasets from $\mathcal{D}$ that are good candidates for further transformation, through dataset retrieval and reranking.
Then, it performs data transformation (i.e., synthetically modify each entries in the selected dataset) to create a new dataset $\mathcal{D'}$ that better aligns with the task requirements.

\subsection{Dataset Retrieval} 

\label{sec:dataset_retrieval} 
The first phase of DataTune involves identifying relevant datasets for our task from a large repository of existing datasets. We confine our dataset space $\mathcal{D}$ to the HuggingFace Hub~\citep{Lhoest2021DatasetsAC}, which consists of over 75,000 datasets.

To efficiently complete this task, we use a dual-stage retrieval approach. We first \emph{retrieve} a set of documents using DataFinder \citep{viswanathan-etal-2023-datafinder}, a bi-encoder retriever specifically trained for retrieving datasets from natural language queries using textual descriptions of each dataset.  We then rerank these datasets to increase the likelihood that our selected dataset can be effectively transformed for the target task.
Inspired by \citet{Sun2023IsCG}, who showed that LLMs can effectively rerank documents, we similarly use LLMs for reranking datasets. While we use only descriptions of the dataset to generate our initial candidate datasets, this is often inadequate for localizing to the best dataset for a given task. For instance, for a math-based task which have multiple choice questions (MCQs), the dataset descriptions of both \textit{GSM8K}\footnote{https://huggingface.co/datasets/gsm8k} and \textit{math\_qa}\footnote{https://huggingface.co/datasets/math\_qa} make them valid choices, but it is only when we look at the schema and sample rows of \textit{math\_qa} (which has multiple choice options) do we see that it is a better choice for our task than \textit{GSM8K} (which has open-ended questions). Thus we also provide additional dataset attributes such as its schema and a small sampling of rows, to aid with the reranking. The reranking step concludes with the name (and any versions) of the chosen dataset $D_R$.
We include the reranking prompt we used in Appendix. \ref{Reranking_Prompt}.\\
Additionally, to improve the reliability of dataset chosen, we employ self-consistency decoding~\cite{wang2023selfconsistency}, where we run the reranker multiple times, and choose the most frequently returned dataset. It is also possible for no suitable dataset being found, acknowledging that not every task has a relevant dataset.

\begin{figure*}[!t]
\begin{tcolorbox}[colback=blue!5!white,colframe=blue!75!black,title=Sample Plan]
 1. Extract the "solutions" field from the dataset as this contains the Python code snippets. \\
 2. For each "solutions" entry, identify the primary operation or functionality of the Python code. This may require parsing the code and understanding its logic. \\
 3. Generate a set of multiple-choice descriptions ("choices") for each code snippet. These should include one correct description of what the code does and several incorrect descriptions. The incorrect descriptions can be plausible but should not accurately describe the code's functionality. \\
 4. Format the "input" field by labeling it as "Python code:" followed by the actual code snippet from the "solutions" field. Below the code, list the generated "choices" with the label "choice:" preceding each option. \\
 5. Determine the correct "choice" that accurately describes the code's behavior. This will be the "output" field. \\
 6. Combine the "input" field and the "output" field to create the final data in the required format for the task examples. \\
 7. If a "solutions" entry does not contain a Python code snippet or is not relevant to the task description, ignore the data sample and return null for that entry.
\end{tcolorbox}
\caption{We show an example plan for the task of providing concise descriptions of Python code. The retrieved dataset contains natural language questions and code solutions. The plan then specifies that the transformation must create the correct description, create incorrect descriptions to create an multiple choice dataset, and format changes required to match the target task examples.}
\label{sample_plan}
\end{figure*}

\subsection{Dataset Transformation}
Having selected an appropriate dataset $D_{R}$ from $\mathcal{D}$, we now focus on tailoring it to the task requirements to create $D'$. This transformation process can include modifications ranging from adjusting the input/output format to more closely match that of the few shot examples to more substantial changes like generating new fields that the present dataset may not currently have. 

To accomplish this complex task, we split it into two primary steps. First, the \textit{Planning Module} generate a multi-step  plan illustrating the sequence of transformations needed to convert examples from $D_{R}$ into the desired output. Next, the \textit{Execution Module} executes the plan on each datapoint in $D_{R}$ to create our resulting synthetic training dataset $D'$. 

Empirically, we notice that the Planning Module is more effective when given descriptions of the source dataset ($D_{R}$) and target task ($T$) that are detailed and specific.
As such, we further implement two more scaffolding modules --- a \emph{Schema Selector} that removes irrelevant columns from dataset $D_R$ (providing a clearer source description) and a \emph{Task Expansion Module} that enriches task descriptions $T$ with requirements (giving a better target specification).

Below, we describe the four modules in their execution sequence (examples in \autoref{fig:data_transformation_architecture}).

\paragraph{Task Expansion} \label{sec:task_explanation}

The brevity of task descriptions often hampers the creation of an effective plan, and often the subtleties of the task are more readily grasped through the examination of examples.\footnote{This is akin to requirement elicitation in Software Engineering, where a deeper understanding of requirements is achieved through further analysis~\cite{lamsweerde2009requirements}.} To address this issue, we implemented an intermediate step wherein the LLM is utilized to scrutinize both the task description and provided examples and generate an expanded version of the task description. This enhanced task specification helps in devising a more detailed and actionable plan, encompassing more explicit steps tailored to the task at hand. The task expansion prompt is linked in the Appendix \ref{Task_Expansion_Prompt}.

\paragraph{Schema Selection}
\label{sec:schema_selection}

The schema selection component is designed by instructing LLMs to identify the most pertinent columns within a dataset for a given task. For example, for the task of code comment generation, where we have a dataset of internet code, the \textit{code snippet} column is extremely useful, whereas the \textit{URL} column code would not be useful. To identify these relevant columns, we provide the LLM with detailed information, including the task description $T$, the chosen dataset $D_R$, the names of existing columns in $D_R$, and samples of dataset rows from $D_R$. The LLM is then tasked with identifying which columns are relevant to the specific task $T$. This approach ensures a targeted selection of dataset features that are directly applicable to the task requirements, optimizing the dataset $D_R$ for the intended application. The schema selection prompt is linked in the Appendix \ref{schema_Prompt}.

\paragraph{Planning Module}
\label{sec:planning}
Now, the LLM generates a comprehensive plan to adapt each data point from the retrieved dataset $D_R$ to the task requirements. This plan is a series of concrete, concise steps. These steps may include combining data fields, elaborating on existing fields, or creating new fields derived from the existing data or excluding the data sample altogether if it is irrelevant. The LLM is provided with the expanded task description, along with the optimized dataset, complete with its description and sample rows. A sample plan is shown in \autoref{sample_plan}.
We are only required to create a plan once per task, since the same task-level plan will get executed for each row of the retrieved dataset. The planning module prompt is linked in the Appendix \ref{Planning_Prompt}.

\begin{table*}[t]
\centering
\resizebox{\linewidth}{!}{
\begin{tabular}{l | l | l |l}
\toprule
\textbf{Task Name} & \textbf{Task Category} & \textbf{Abbreviation} & \textbf{Task Instruction} \\
\midrule
Temporal Sequences & Logical Reasoning & Time & Answer questions about which times certain events could have occurred.\\
\midrule
Code Line Descriptions & Coding & Code & Give an English language description of Python code.\\
\midrule
Elementary Math & Math & Math & Answer a multiple choice mathematical word problem.\\
\midrule
Cause and Effect & Causal Reasoning & C\&E & Answer multiple-choice questions distinguishing cause and effect.\\
\midrule
Medical Questions in Russian & Domain Specific & Russian & Answer a yes/no question about medical text in Russian.\\
\midrule
Implicatures & Contextual QA & Impl. & Predict whether Speaker 2's answer to Speaker 1 is affirmative or negative.\\
\bottomrule
\end{tabular}
}
\caption{We evaluate our method on 6 diverse text-based tasks from BIG-Bench.}
\label{tab:task_abbreviations}
\end{table*}

\paragraph{Execution Module}
\label{sec:execution}
The execution of the transformation plan for each dataset sample from $D_R$ is done using an LLM. The LLM receives three key pieces of information for each row of the dataset: the row itself, detailed specifications of the input task, and the transformation plan formulated earlier. Conditioned on these pieces of information, the model responds by producing an adjusted dataset row which, hopefully, meets the requirements of the input task. The execution module prompt is linked in Appendix \ref{Transform_Prompt}. The adjusted dataset rows form our synthetic dataset $D'$ used for fine-tuning.

\subsection{Using Multiple Datasets} 
The transformation process of data points may result in a significant number of them getting filtered out. Additionally many datasets may be small to begin with. Both these factors can result in a reduced quantity of transformed data. 
To balance quantity and quality, we adopt a strategy of transforming multiple highly ranked datasets until we reach our desired dataset set.
Furthermore, if a considerable proportion of data sample transformations fail (e.g. noisy data samples in the dataset) for a specific dataset, we opt to exclude it from consideration.

\section{Experimental Setup}

\paragraph{Evaluation procedure} We evaluate DataTune and related baselines over 6 tasks from the BIG-Bench benchmark \cite{srivastava2023beyond}, which we list in \autoref{tab:task_abbreviations}. BIG-Bench focuses on tasks believed to be beyond the capabilities of current language models, covering a variety of task categories. We choose tasks spanning logical reasoning, coding, math, causal reasoning, multilinguality, and domain specific tasks. For each task, we create 3000 data points through DataTune, and apply the training procedure highlighted earlier. We evaluate all models for two-shot performance, as per the BIG-Bench testing suite.\footnote{\href{https://github.com/google/BIG-bench}{https://github.com/google/BIG-bench}}\\

\paragraph{Methods} 
We compare the effectiveness of various \textbf{dataset collection} strategies, by using the resulting data to finetune the same base model, Mistral-7B~\cite{jiang2023mistral}, a leading open-source LLM within its size category. As shown in \swedit{\autoref{tab:performance_comparison}}, this includes:
(1) retrieving existing data using dense retrieval \cite{viswanathan-etal-2023-datafinder}, (2) generating synthetic data, (3) Prompt2Model~\cite{prompt2model} (a state-of-the-art method that combines retrieving existing data and synthetic data generation), (4) our DataTune approach, and (5) a combination of DataTune and synthetic data generation, which represents an integration of all the existing methods.

We also include two \textbf{prompting} baselines: (6) 
few-shot prompting on the base model Mistral-7B, and (7) \texttt{GPT-3.5-Turbo}, a significantly larger model, which we include as a robust benchmark due to its extensive capabilities.

\paragraph{Dataset Creation Setup} Our initial retrieval step using DataFinder retrieves 25 candidate datasets that are processed for reranking. We transform upto 4 datasets per task until we meet our desired set of 3000 data points. The LLM used for all components is GPT-4-Turbo, except for the final execution step, which uses GPT 3.5-Turbo.\footnote{Our dataset transformation method requires making an LLM query for each instance in the dataset. Therefore, the decision to use GPT-3.5-turbo was made for budgetary reasons.}

\paragraph{Training Setup} We used the Mistral-7B model, following the approach of \citet{jiang2023mistral}, and applied the QLoRA technique from \citet{dettmers2023qlora} for fine-tuning. The process was carried out over 3 epochs. We select parameters by running 4 runs across two sets of hyperparameters over two values( learning rate: {$5e^{-5}$ and $1e^{-4}$}, and QLoRA's $\alpha$ parameter between 16 and 24 and choose the run with the lowest validation loss at any point. We used the AdamW optimizer \cite{DBLP:journals/corr/abs-1711-05101} and set QLoRA $r=8$. We conduct our training on 2 NVIDIA RTX A6000 GPUs.\\

\paragraph{Metrics} Following \citet{srivastava2023beyond}, we use a \textit{normalized aggregate score} evaluation that normalizes a multiple choice grade score, such that a score of 0 implies random chance performance, and 100 implies human expert performance. Additionally, a score less than 0 indicates performance worse than random chance.

\newcommand{\rotheader}[1]{\rotatebox{90}{#1}\hspace{-30pt}}
\begin{table*}[] 
    \centering
    \small
    \resizebox{.9999\linewidth}{!}{
    \begin{tabular}{c  r   | ll  | l | r r  r  r  r  r | r }
    \toprule
    \multirow{1}{{0.03\textwidth}}{} & \multicolumn{1}{c}{\textbf{Method}}
    & \multicolumn{2}{|c|}{\textbf{Steps}} & \multirow{2}{{0.07\textwidth}}{\centering \textbf{\# Train.\\Points}}
    & \multicolumn{7}{c}{\textbf{Tasks}} \\

    \cmidrule(r){3-4} \cmidrule(l){6-12} 
    \multicolumn{2}{p{0.03\textwidth}|}{} 
    & Retrieval Type
      & Generation
      & 
      & Time
      & Code
      & Math
      & C\&E
      & Russn.
      & Impl.
      & \textbf{Avg.} \\
    \midrule

    & \multicolumn{3}{c}{\textbf{Few-Shot Baselines}} & \multicolumn{7}{c}{} \\
    \multirow{2}{{0.03\textwidth}}[-0.5em]{}
    & GPT-3.5
        & - & - & 0 
        & 50.6 & 75.6 & 30.4 & 96.7  &  90.6 & 64.2 & 68.0\\
    & Mistral-7B
        & - & - & 0
        & -2.5 &  62.3 & 2.9 & 37.2 & 39.8 & 39.0 & 29.8\\
    \midrule

    \multirow{6}{{0.03\textwidth}}[0pt]{\rotheader{Mistral-7B+}} & \multicolumn{3}{c}{\textbf{Synthetic Finetuning Baselines}} & \multicolumn{7}{c}{} \\

    & Existing data 
        & Dense & - & 3000
        & -4.7 & 62.3 & 0.8 & 52.9 & 0.0 & 39.9 & 25.2 \\
    & Synthetic data 
        & - & Synthetic & 3000
        &  2.0 & 60.8 & 3.8 & 37.2 & 54.0 & 41.9 & 33.3 \\
    & DataTune (DT) 
        & + Reranker & Transformed  & 3000 & -2.1 & 71.2 & 1.3 & \textbf{56.9 }& 48.0 & 41.9 & 36.2 \\
    & Prompt2Model 
        & Dense & Synthetic
        & 6000 & -2.0 & 73.4 & 4.7 & 33.8 & \textbf{86.0} & 44.0 & 40.0 \\
    & DT+Synthetic 
        & + Reranker & Both 
         & 6000 & \textbf{16.9}  & \textbf{84.5} & \textbf{8.1} & 41.2 & 68.0 & \textbf{48.0} & \textbf{44.5}\\

    \bottomrule
    \end{tabular}
    }
    \caption{We compare the performance of different few-shot learning methods across six BIG-Bench tasks. Here, we categorize each method by what base model is used (Mistral-7B or GPT-3.5-Turbo), whether data is retrieved (and, if so, whether a dense retriever or dense retriever + reranker is used), how the data points are generated (whether transformed from an existing dataset, generated synthetically, or both). ``\# Train. Points'' refers to the number of training examples produced by the method for each task. Normalized aggregate scores below zero imply performance worse than chance. }
    \label{tab:performance_comparison}
\end{table*}

\section{Results and Analysis}

\subsection{Performance Comparison}

\emph{DataTune consistently outperforms few-shot prompting, as well as existing individual data collection methods.}
From Table \ref{tab:performance_comparison}, we see that fine-tuning our base model on DataTune outperforms the base Mistral-7B model by 6.4 points on average, improving over it in five out of six tasks.
We also show that DataTune provides an average improvement of 11 and 2.9 points over fine-tuning on  existing data and synthetically generated data respectively.

\emph{DataTune's transformation is complementary to synthetic data generation.}
It is noteworthy that the combination of DataTune and synthetically generated data results in a marked performance increase. This synergistic improvement yields an overall average score of 44.5 on the BIG-Bench tasks we consider. We provide a more detailed analysis of the synergy between DataTune and existing synthetic dataset generation methods in Section \ref{sec:complementary}.

\emph{Our system (DataTune + Synthetic Data Generation) outperforms SOTA baselines like Prompt2Model.}
In order to make a fair comparison with baselines such as Prompt2Model that fine-tune on both existing data and synthetically generated data, we define our system as the base Mistral-7B model fine-tuned on DataTune-created data and synthetically generated data.
Our analysis in Table \ref{tab:performance_comparison} highlights significant differences in performance between our system and Prompt2Model, with our system demonstrating a notable advantage over Prompt2Model five out of six tasks, with an average improvement of 8.3 points. These findings underscore the effectiveness of our system to create quality datasets for fine-tuning across a wide range of tasks and domains.

\subsection{DataTune Impact on Dataset Diversity}
\label{sec:dataset_diversity}
Sampling diverse yet high-quality examples from a language model is a challenging task. Prior work has shown that the correctness of a synthetic dataset sampled from a language model is inversely correlated with its diversity \citep{Ye2022ProGenPZ}. Generating synthetic data directly from a language model (using a method like Prompt2Model) often contains near-duplicate examples.

\begin{figure}[t]
\includegraphics[width=0.48\textwidth]{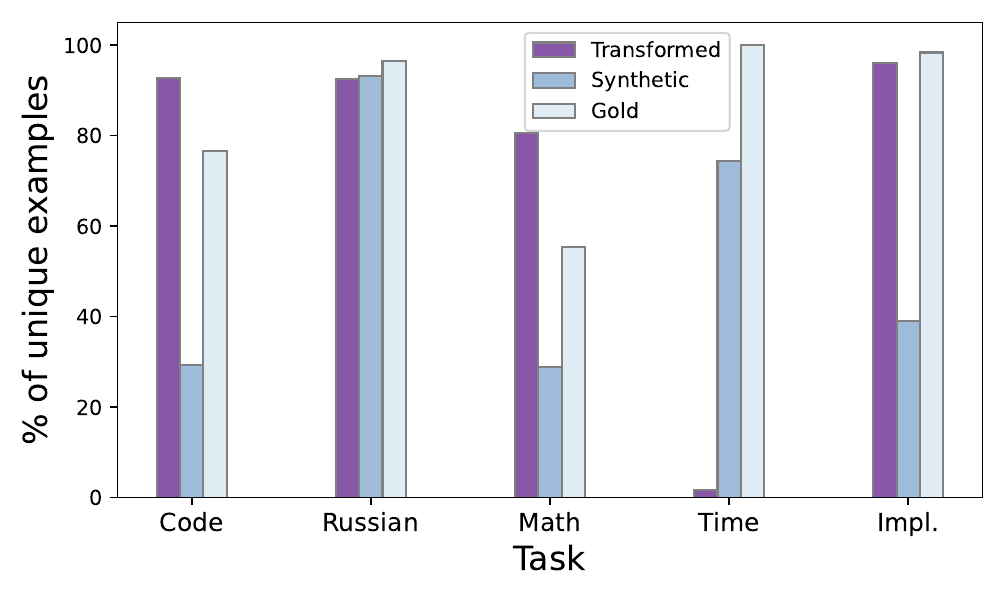}
\caption{Synthetic dataset generation often suffers from the problem of generating multiple duplicates of the same example in a given dataset. On 3 of 5 tasks, we find that data transformation from retrieved datasets significantly mitigates this issue. The other two datasets, Russian and Temporal, represent failure modes of our system. Gold represents the BigBench Dataset for a given task.}
\label{fig:deduplication}
\end{figure}

Does transforming existing datasets reduce the incidence of duplicate examples? Using ROUGE-L to determine lexical uniqueness \citep{rouge, wang2022self}, we determine a sentence $s$ in dataset $D'$ to be \textit{unique} up to threshold\footnote{We set the ROUGE-L threshold to 0.8 for Code Line Descriptions, where examples are Python snippets, 0.9 for Temporal Sequences, where examples are long English texts, and 0.7 for the other datasets.} $T$ if $ \max_{s' \in D' \setminus \{s\}} ROUGE(s, s') < T$. In \autoref{fig:deduplication}, we observe that over 50\% of examples generated synthetically are near-duplicates for 3 of 5 tasks; in each of those cases, using DataTune instead effectively eliminates this problem.

We observe similar trends with lexical diversity across all dataset creation methods. In \autoref{table:lexical_diversity}, we measure the number of unique bigrams per generated input example. DataTune significantly increases the number unique bigrams per example (and moderately increases the length of each example) on 3 of 5 datasets.

According to both measures, dataset diversity decreases on \textit{Temporal Sequences} and \textit{Medical Questions in Russian}, which are also the two tasks where DataTune fails to improve over training on fully synthetic data. We discuss these two tasks in Section \ref{sec:complementary} and \hyperref[sec:limitations]{Limitations}, respectively.

\begin{table}[t] 
  \centering
  \renewcommand{\arraystretch}{1.1}
  \fontsize{9.5}{10}\selectfont
  \setlength{\tabcolsep}{2.5pt}
  \begin{tabular}{@{} r r r @{} }
  \toprule
  \multicolumn{1}{l}{\textbf{Dataset}} & \multicolumn{1}{l}{\textbf{Unique Bigrams}} & \multicolumn{1}{l}{\textbf{Total Tokens}} \\
  \multicolumn{1}{l}{} & \multicolumn{1}{l}{\textbf{Per Example}}  & \multicolumn{1}{l}{\textbf{Per Example}} \\
  \midrule
  \multicolumn{3}{c}{\textbf{Code Line Description}}
  \\
Gold   & 13.2 & 32.3  \\
  Synthetic & 2.5 & 35.0 \\
  Transformed  & 14.9 & 86.9   \\
  \midrule
  \multicolumn{3}{c}{\textbf{Elementary Math}}
  \\
Gold   & 10.8 & 48.6   \\
  Synthetic  & 3.3 & 34.4 \\
  Transformed  & 11.6 & 43.8   \\
  \midrule
  \multicolumn{3}{c}{\textbf{Implicatures}}
  \\
Gold  & 9.9 & 24.1   \\
  Synthetic & 2.7 & 27.7 \\
  Transformed  & 17.8 & 39.8  \\
  \midrule
      \multicolumn{3}{c}{\textbf{Temporal Sequences}}
  \\
Gold  & 1.0 & 99.7   \\
  Synthetic & 20.8 & 54.6 \\
  Transformed  & 0.2 & 73.7  \\
  \midrule
  \multicolumn{3}{c}{\textbf{Medical Questions in Russian}}
  \\
Gold   & 62.0 & 79.4   \\
  Synthetic & 20.8 & 54.6 \\
  Transformed  & 11.6 & 44.8  \\
  \bottomrule
  \end{tabular}
  \caption{We observe that dataset transformation yields datasets with greater lexical diversity than synthetic dataset generation on 3 of 5 datasets.
  }
  \label{table:lexical_diversity} 
\end{table}

\subsection{DataTune Generates Harder Examples}
\label{sec:hard_example}
\begin{figure}[t]
\includegraphics[width=0.48\textwidth]{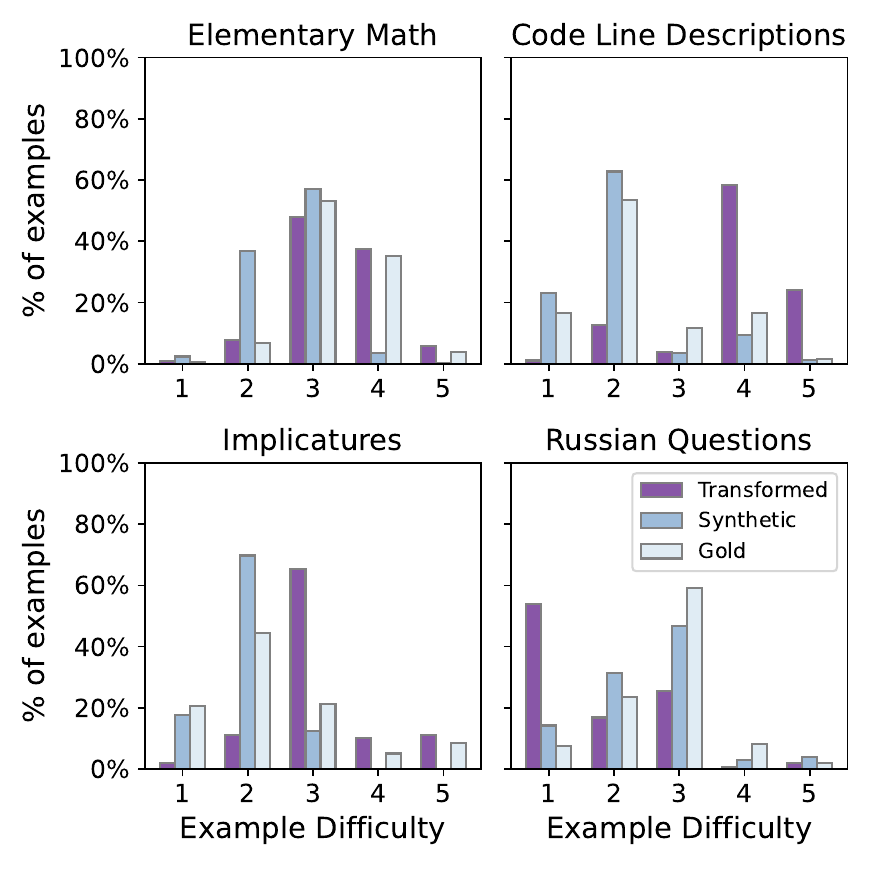}
\caption{Dataset Transformation leads to more difficult examples than synthetically generated examples, which are over-represented by easy examples, relative to manually-curated BIG-Bench evaluation datasets (Gold).}
\label{Dataset_Difficulties}
\end{figure}

Synthetic datasets sampled directly from a language model tend to overrepresent with easy-to-solve examples \citep{xu2023wizardlm}. To test our ability to overcome this issue, we used \texttt{gpt-3.5-turbo} to estimate the difficulty of each example generated from each method (DataTune, synthetic generation, and manual curation) for four tasks. We wrote a custom difficulty estimation prompt (including in-context examples) for each task; in all prompts, we specify that the LLM should rate the example on a scale of 1 to 5. We outline an example of this prompt in Appendix \ref{difficulty_prompt}.  As illustrated in \autoref{Dataset_Difficulties}, we observe that the data from DataTune exhibits a higher level of difficulty compared to synthetically generated data for all datasets other than \textit{Medical Questions in Russian}.  We attribute this increase in difficulty to the added knowledge introduced by the existing dataset selected for transformation, which is often more nuanced and, consequently, more challenging. In contrast, unconditional generation from LLMs tends to produce examples that are likely to have a high prior (prompt-independent) probability under the language model. The exception to this effect, \textit{Medical Question in Russian}, can be attributed to the failure of our approach's transformation plan to generate data in the correct language for this task. We provide more details of this failure mode in \hyperref[sec:limitations]{Limitations}.

\subsection{Does generating harder examples lead to lower label accuracy?}
\label{sec:label_accuracy}
We evaluated the label accuracy of generated data for the Code Line Description task by comparing DataTune against synthetic dataset generation. Manually annotating a random sample of 300 data points from each generation method for the Code Line Description Task, we found the label accuracy from DataTune is 88\%, compared to 86.6\% for synthetic data generation. This comparison suggests that DataTune can produce datasets that are comparable in accuracy to purely-synthetic datasets, despite DataTune generating significantly more diverse and challenging examples on this task.

\subsection{Transformed Data Can Be Complementary to Synthetic Data}
\label{sec:complementary}

In two cases where DataTune fails to improve over a synthetic data baseline, \textit{Temporal Sequences} and \textit{Elementary Math}, our combined DataTune + Synthetic system still outperforms all other comparable (Mistral-based) baselines. We observe these two approaches to dataset generation can be complementary to each other, yielding additive improvements when combined.

\begin{figure}[t]
\includegraphics[width=0.48\textwidth]{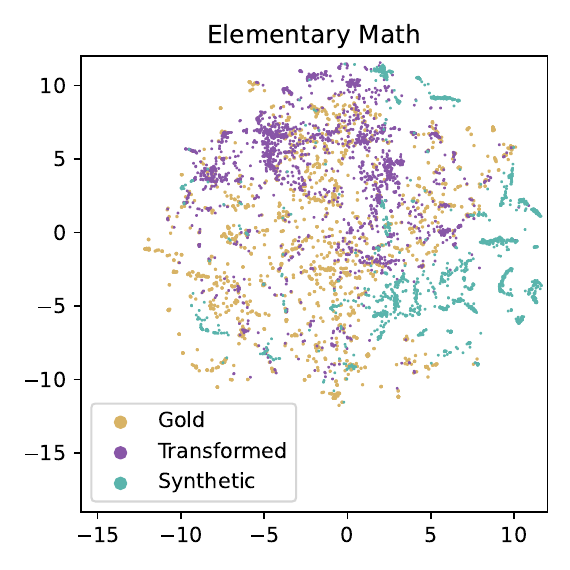}
\caption{We observe that examples generated via DataTune and synthetic dataset generation fall into visually well-separated regions of embedding space, after projecting to two dimensions via t-SNE.}
\label{fig:elementary_math_embeddings}
\end{figure}

For a concrete example of this, we can visually observe the two-dimensional semantic distribution of questions generated for the Elementary Math task via DataTune, Synthetic Generation, and from the gold dataset in \autoref{fig:elementary_math_embeddings}. We encoded each question using MiniLM v2 \citep{minilm} via \texttt{sentence-transformers} \citep{reimers-2019-sentence-bert}, then projected each embedding into a 2D space using t-SNE \citep{tsne}.

Visually, we observe that the embedding clusters of questions generated via DataTune and Synthetic Generation appear to be largely disjoint. This supports our hypothesis that these two methods systematically cover different distributions of the target task space, and therefore combining the examples from each method can lead to a synergistic effect.

\section{Conclusion and Future Work}
 We presented DataTune, an improved approach for enhancing automatic dataset generation by transforming existing labeled datasets. Our method significantly outperforms existing automatic dataset generation techniques on several challenging tasks from BIG-Bench. When used in conjunction with existing synthetic dataset generation methods, it achieve superior performance compared to other few-shot learning methods on the base model we used (Mistral-7B). Our analysis reveals that DataTune not only creates more diverse and accurate datasets but also increases their complexity for fine-tuning. An important direction is whether transformation-based dataset creation methods like DataTune can still be effective when examples are retrieved from the open web rather than from the collections of manually-curated datasets we consider in our work. Another important direction for future work will be to generate code to execute a transformation plan (rather than querying an LLM for each instance of data). Both of these future directions would improve the accessibility and scalability of our suggested approach.

\section*{Limitations}
\label{sec:limitations}
We identify four key limitations in our system:\\ 
\begin{enumerate}

\item \textbf{LLM Query Cost}: Our dataset transformation component requires running the Execution Module on each row of each dataset we wish to transform. Given that our Execution Module prompts a large language model, the number of LLM queries scales linearly with the amount of data to be transformed. This could be cost-prohibitive for transforming very large datasets. This LLM usage requirement could also exclude members of the research community without consistent LLM access from benefiting from our work.

\item \textbf{Dependence on the Planning Module}: Our dataset transformation system relies heavily on the Planning Module to produce clear and comprehensive instructions for the Execution Module to enact. Our Planning Module operates by prompting a large language model. Given that prompted LLMs can behave unpredictably in changes in prompts \citep{sclar2023quantifying}, an important limitation of our method is that its success depends on a large language model following a prompt correctly. We see this failure with Medical Questions in Russian, where the agent is supposed to translate a Russian dataset into English to facilitate the generation of question-answer pairs. When running our benchmarking, the Planning Agent failed to translate the generated question-answer pairs back to Russian. This resulted in a training dataset that, despite being conceptually close, was practically far from correct.

\item \textbf{Handling Non-English Data}: Our transformation agent's ability to process non-English data tasks is substantially compromised, frequently altering example tasks instead of the actual data. This deficiency is primarily due to the reliance on models like GPT 3.5, which have been extensively trained on English data, thereby diminishing their proficiency with other languages.

\item \textbf{Dependence on Instruction-Following LLMs}: The system's execution component depends on Large Language Models (LLMs) that are specifically designed to adhere to instructions. We have identified discrepancies in performance among LLMs tailored for instruction-based tasks versus those developed for conversational purposes. This limitation confines our system to using only a narrow selection of LLMs that demonstrate the ability to follow instructions accurately.
\end{enumerate}

\section*{Ethical Considerations}

DataTune could make it easier for the general public to build custom language models. The broadening of open-ended technology  induces ethical concerns, similar to the issues with open-source deepfake libraries described by \citet{10.1145/3531146.3533779}. While DataTune has potential for misuse, this is likely no greater than the potential harms presented by the underlying open-source large language models. By making it easier to build task-specific language models, we hope that these risks are balanced by the benefits of making NLP models accessible to those outside the NLP community or those without the resources to manually collect labeled data. We aim to be transparent in our documentation about the potential limitations of the system. 

The platform is particularly valuable for individuals outside the NLP community who can benefit from using NLP models in their work but lack the specialized knowledge to develop these tools independently. The decision to open-source DataTune invites community contributions, emphasizing a collaborative approach to improve and expand the tool's capabilities. This strategy not only enhances the system's utility but also aligns with a broader goal of increasing the accessibility of NLP innovations and fostering a more inclusive technological environment.

\bibliography{anthology,custom}
\bibliographystyle{acl_natbib}
\onecolumn
\appendix

\section{Example Appendix}
\label{sec:appendix} 

\subsection{Reranking Prompt}
\label{Reranking_Prompt}

\begin{tcolorbox}

Your objective is to choose the most relevant dataset for a given a task (and few examples of the task). For each dataset, you will be provided with the dataset description, and tags related to the dataset. Please return the most relevant dataset, e.g., squad 

The following is the task \\ \{\{instruction\}\} \\ and these are some examples of the same: \\\{\{examples\}\} \\
There are \{\{num\}\} datasets available for this task. \\
\{\{datasets\}\}\\
The name of the most relevant dataset for this task is:
\end{tcolorbox}

where each dataset in \{\{datasets\}\} is defined as: 
\begin{tcolorbox}
\{counter\}\{\{dataset\_name\}\}:Description-\{\{dataset\_description\}\}.\\This dataset has the following tags:\\ \{\{tags\}\} 
\end{tcolorbox}

\subsection{Schema Selection Prompt}
\label{schema_Prompt}
\begin{tcolorbox}
Your objective is to carefully analyze the task and the dataset mentioned, and decide whether the columns are relevant input, relevant output, irrelevant for the given task, or if it is ambiguous. There should be at most one output column. It is possible to have no relevant columns, in which case return the input and output column as empty lists.  Answer in a json format, with the following keys: input, output, irrelevant, ambiguous.\\
\{\{INCONTEXT\_EXAMPLES\}\}\\
After seeing these examples with the required columns, please provide the relevant columns for this context:
\\\\
You are tasked with the following process. \{\{instruction\}\} For this task, you will use the \{\{dataset\_name\}\} dataset from HuggingFace. Dataset Description: \{\{dataset\_description\}\} \\A sample data instance from this is as follows. \{\{sample\_row\}\}.\\This dataset has the following columns: \{\{dataset\_columns\}\}\\ 
Required Columns :
\end{tcolorbox}
\subsection{Planning Module Prompt} 
\label{Planning_Prompt}
\begin{tcolorbox}
    
You are a Planning Agent. You create a plan to transform data samples from their existing format into the required format for a given task.

-------------------------------------------------\\
Here are some examples for your reference.

\{\{in\_context\_examples\}\}

------------------------------------------------\\
Now do the following task: 

Task Description: \{\{task\_description\}\}

Task Examples:
{example}

Here are samples from a potentially relevant dataset for the task above. Notice how the format below is not as required by the task above.

Dataset Samples: 
\{\{dataset\_row\}\}

Carefully analyze the  `Task Description` and the `Task Examples`. Propose a higher-level plan to convert data from the Dataset Sample to data in the required format task examples. Your plan should be a list of sequential steps that can be taken to perform the data transformation. You don't need to use all columns, as the dataset may not be fully relevant. Keep steps as simple, explicit and concise as possible. Each step in the plan may take any of the following actions:
1. Generate new columns as required by the task, and save them \\
2. Expand on a particular column to make it something more relevant to the task and save it \\
3. Combine multiple columns from the dataset \\
4. Choose columns that will form "input" \\
5. After the input field is created, carefully analyze it to choose/generate the output field \\
6. Ignore a data sample because it is not all relevant and return null for them. \\

Return only the plan.
\end{tcolorbox}

\subsection{Execution Module Prompt}
\label{Transform_Prompt}
\begin{tcolorbox}
    
You are a Data Transforming Agent. Your job is to transform data from a given format to the required format. Following are the detailed instructions for the same:
1. Read the `Task Description`. \\
2. An example of the input and output looks like for the task is shown in `Task Examples`\\
3. The sample to be transformed is in `Data Sample`. \\
4. Read the data transformation plan carefully that will help you convert the `Data Sample` into the required format. This should be relevant and intune to the `Task Description` \\
5. Perform the plan step by step and explain your thinking. \\
6. End your response with the transformed sample as a JSON response with exactly 2 fields: "input" and "output". 
\\-------------------------------------------------\\
Here are some examples for your reference.
\{\{incontext\_examples\}\}
\\------------------------------------------------\\
Now do the following task: 

Task Description: \{\{task\_description\}\}

Task Examples:
\{\{sample\}\}

\{\{plan\}\}

Dataset Sample:
\{\{dataset\_row\}\}

Think step by step through the plan to convert the above `Dataset Sample` and show your working. End your response as a JSON with exactly two fields: "input", and "output"
Response:
\end{tcolorbox}

\subsection{Task Expansion Prompt}
\label{Task_Expansion_Prompt}
\begin{tcolorbox}
    
Carefully analyse the  task description and examples of the task, and explain the task to give a clearer description. Do not explain each example, but rather capture the general trends. Also place special focus on the format of the input/output examples. 
\\-------------------------------------------------\\

Task Description: \{task description\}

Task Examples: \{examples\}
\end{tcolorbox}
\subsection{Difficulty estimation prompt: Code Line Descriptions example}
\label{difficulty_prompt}
\label{Task_Expansion_Prompt}
\begin{tcolorbox}
\begin{verbatim}
We are building a dataset for automatically describing code
(in words). Evaluate and rate the difficulty and complexity
of describing the following code lines. You should give an
overall score on a scale of 1 to 5,
where a higher score indicates higher difficulty.
You must just give a score without any other reasons.
Here's the grading scale:
1: Very easy. Anyone who understands the 
programming language could describe this almost instantly
2: Easy. Anyone who understands the programming 
language could describe this with a bit of thought
3. Neutral. Most non-expert people who understand 
the programming language would be able to describe this,
but it might take time for them to understand the code
4. Hard. It would require at least a minute 
for a non-expert person who understand the
programming lanugage to understand and describe this code.
5. Very hard. Most non-experts would make a mistake 
when trying to describe this code in a fixed timeframe.
Professional programmers would have an easier time.

Your answer shoud be a single number, 1 through 5, 
with nothing else in your response.

{Incontext examples with code and difficulty}

{Input Code}
\end{verbatim}
\end{tcolorbox}

\end{document}